\documentclass{article}

\usepackage{PRIMEarxiv}
\usepackage{longtable}
\usepackage{ltablex}
\usepackage[utf8]{inputenc} 
\usepackage[T1]{fontenc}    
\usepackage{hyperref}       
\usepackage{url}            
\usepackage{booktabs}       
\usepackage{amsfonts}       
\usepackage{nicefrac}       
\usepackage{microtype}      
\usepackage{lipsum}
\usepackage{graphicx}
\usepackage{tabularx}
\usepackage{multirow}
\usepackage[export]{adjustbox} 
\graphicspath{{media/}}     
\usepackage{makecell}
\usepackage{subcaption}
\keepXColumns
  
\title{The Great March 100: 100 Detail-oriented Tasks for Evaluating Embodied AI Agents}

\author{
Ziyu Wang$^1$, Chenyuan Liu$^1$, Yushun Xiang$^1$, Runhao Zhang$^2$, \textbf{Yu Zhang}$^1$, \textbf{Qingbo Hao}$^1$, \textbf{Hongliang Lu}$^1$\\
\textbf{Houyu Chen}$^1$, \textbf{Zhizhong Feng}$^1$, \textbf{Kaiyue Zheng}$^1$, \textbf{Dehao Ye}$^1$, \textbf{Xianchao Zeng}$^2$, \textbf{Xinyu Zhou}$^2$, \textbf{Boran Wen}$^{1,2}$\\
\textbf{Jiaxin Li}$^{1,2}$, \textbf{Mingyu Zhang}$^{1,2}$, \textbf{Kecheng Zheng}$^3$, \textbf{Qian Zhu}$^3$, \textbf{Ran Cheng}$^3$, \textbf{Yong-Lu Li}$^{1,2}$\thanks{Corresponding author.} \\
  $^1$SJTU, $^2$SII, $^3$Robbyant\\
  \texttt{\{ziyu.wang, yonglu\_li\}@sjtu.edu.cn},\\
  \texttt{zkechengzk@gmail.com, \{yunzhong.zq, zhouzhan.cr\}@antgroup.com} \\
  RHOS.ai, xbench
  }

\begin{document}
\maketitle

\begin{abstract}
Recently, with the rapid development of robot learning and imitation learning, numerous datasets and methods have emerged. However, these datasets and their task designs often lack systematic consideration and principles. This raises important questions: 
Do the current datasets and task designs truly advance the capabilities of robotic agents? 
Do evaluations on a few common tasks accurately reflect the differentiated performance of various methods proposed by different teams and evaluated on different tasks? 
To address these issues, we introduce the Great March 100 (\textbf{GM-100}) as the first step towards a \textbf{robot learning Olympics}. GM-100 consists of 100 carefully designed tasks that cover a wide range of interactions and long-tail behaviors, aiming to provide a diverse and challenging set of tasks to comprehensively evaluate the capabilities of robotic agents and promote diversity and complexity in robot dataset task designs. These tasks are developed through systematic analysis and expansion of existing task designs, combined with insights from \textbf{human-object interaction primitives and object affordances}. We collect a large amount of trajectory data on different robotic platforms and evaluate several baseline models. Experimental results demonstrate that the GM-100 tasks are 1) feasible to execute and 2) sufficiently challenging to effectively differentiate the performance of current VLA models.
Our data and code are available at \url{https://rhos.ai/research/gm-100}.

\end{abstract}

\section{Introduction}
Recently, with the rapid development of robot learning, numerous datasets and task designs were proposed. For example, Open X-Embodiment~\cite{embodimentcollaboration2025openxembodimentroboticlearning} assembled a dataset from 22 different robots, containing 160,266 tasks. Agibot~\cite{bu2025agibot} collects 200+ tasks with 1M+ trajectories; RoboCOIN \cite{RoboCOINReport} collects over 180,000 demonstrations for 421 tasks. However, these datasets and tasks often focus on a few common tasks and behaviors. After removing duplicates and categorizing them based on their semantic meanings, most tasks concentrate on very common behaviors such as ``pick and hold'', while lacking coverage of complex and long-tail tasks. This singular task design leads to significant biases in the trained models, limiting their applicability in real-world scenarios as \textit{pre-trained models}, except for a few common tasks.
Similarly, current evaluation tasks suffer from analogous issues. Most studies, when proposing new methods, tend to test only on a few common tasks, without a unified task design standard, making fair comparisons across different works difficult.  

To address these issues, we introduce the Great March 100 (\textbf{GM-100}) as the first step towards a robot learning Olympics.
GM-100 consists of 100 carefully designed tasks that cover a wide range of interactions and long-tail behaviors, aiming to provide a diverse and challenging set of tasks to comprehensively evaluate the capabilities of robotic agents and promote diversity and complexity in robot dataset task designs. These tasks are developed through systematic analysis and expansion of existing task designs, combined with insights from human action understanding. We collect a large amount of trajectory data on two different robotic platforms and evaluate several baseline models. Experimental results demonstrate that the GM-100 tasks are 1) feasible to execute and 2) sufficiently challenging to effectively differentiate the performance of various methods.

Besides, in the task design process, we do not rely on the \textit{utility} for real-world tasks as the standard to avoid human bias, but follow the \textbf{physical common sense} and \textbf{low-level manipulation knowledge (the how-level affordance)} as the only standards to generate and select the final tasks. 

To summarize, in this report, we make the following contributions:
\begin{itemize}
    \item We identify the limitations of existing robot task designs and evaluations, highlighting the need for more diverse and complex tasks.
    \item We propose GM-100, a task list consisting of 100 detail-oriented tasks that cover a wide range of interactions and long-tail behaviors.
    \item We collect a medium-sized dataset on robotic platforms and evaluate several baseline models, demonstrating the challenge and effectiveness of GM-100.
\end{itemize}

Our data and code are available at \url{https://rhos.ai/research/gm-100}.

\section{Related Work}
\label{sec:related}
\subsection{Imitation Learning}
Imitation learning underpins embodied intelligence by teaching agents to map sensory inputs to actions via expert demonstrations. Early methods include Behavioural Cloning \cite{Pomerleau-1989-15721}, interactive aggregation as in DAgger \cite{ross2011reduction}, adversarial approaches like GAIL \cite{ho2016generative}. More recently, diffusion-based policies such as ACT \cite{zhao2023learning}, Diffusion Policy \cite{chi2023diffusion}, and 3D Diffusion Policy \cite{ze20243d}. These techniques improve sample efficiency and multimodal trajectory modeling but still face challenges in distributional shift, real-time inference, and training stability.

\subsection{Vision-Language Action Model}
Building on these foundations, Vision–Language–Action (VLA) models merge perception, instruction understanding, and control into unified networks. Representative instances include RT-2 \cite{zitkovich2023rt}, OpenVLA \cite{kim2024openvla}, Robotics Diffusion Transformer (RDT) \cite{liu2024rdt}, $\pi_{0}$ \cite{black2024pi_0}, CogACT \cite{li2024cogact}, SpatialVLA \cite{qu2025spatialvla}, $\pi_{0.5}$ \cite{intelligence2025pi_}, SmolVLA \cite{shukor2025smolvla}, UniVLA \cite{bu2025univla}, WALL-OSS \cite{zhai2025igniting}, GR00T \cite{GR00Tn1_2025}, RynnVLA-002 \cite{cen2025rynnvla}. Despite their effectiveness, both traditional and VLA-based imitation learning methods often require large-scale expert demonstrations.

\subsection{Manipulation Datasets and Task Design}
To advance the learning and training of robotic policies, numerous datasets have emerged in recent years, such as Open X-Embodiment \cite{embodimentcollaboration2025openxembodimentroboticlearning}, Agibot \cite{bu2025agibot}, BridgeData V2 \cite{walke2023bridgedata}, RH20T \cite{fang2024rh20t}, DROID \cite{khazatsky2025droidlargescaleinthewildrobot}, RoboCOIN \cite{wu2025robocoinopensourcedbimanualrobotic}, RoboMIND \cite{wu2025robomind}. 
However, while there has been a surge in data collection, few have focused on task design and balancing task diversity and long-tail representation. This has led to significant overlap in datasets and task designs, concentrating on a few common tasks and behaviors like ``pick and hold'' while lacking coverage of complex and long-tail tasks. This limitation hinders the development of intelligent agents with truly human-like capabilities. 
Similarly, evaluations of robotic agents often focus on these common tasks, neglecting complex and long-tail tasks, which limits our comprehensive assessment of agent capabilities.

\section{Task Design of GM-100}
\label{sec:task}
\begin{figure}[t]
  \centering
  \begin{minipage}[c]{0.45\linewidth} 
    \centering
    \includegraphics[width=\linewidth]{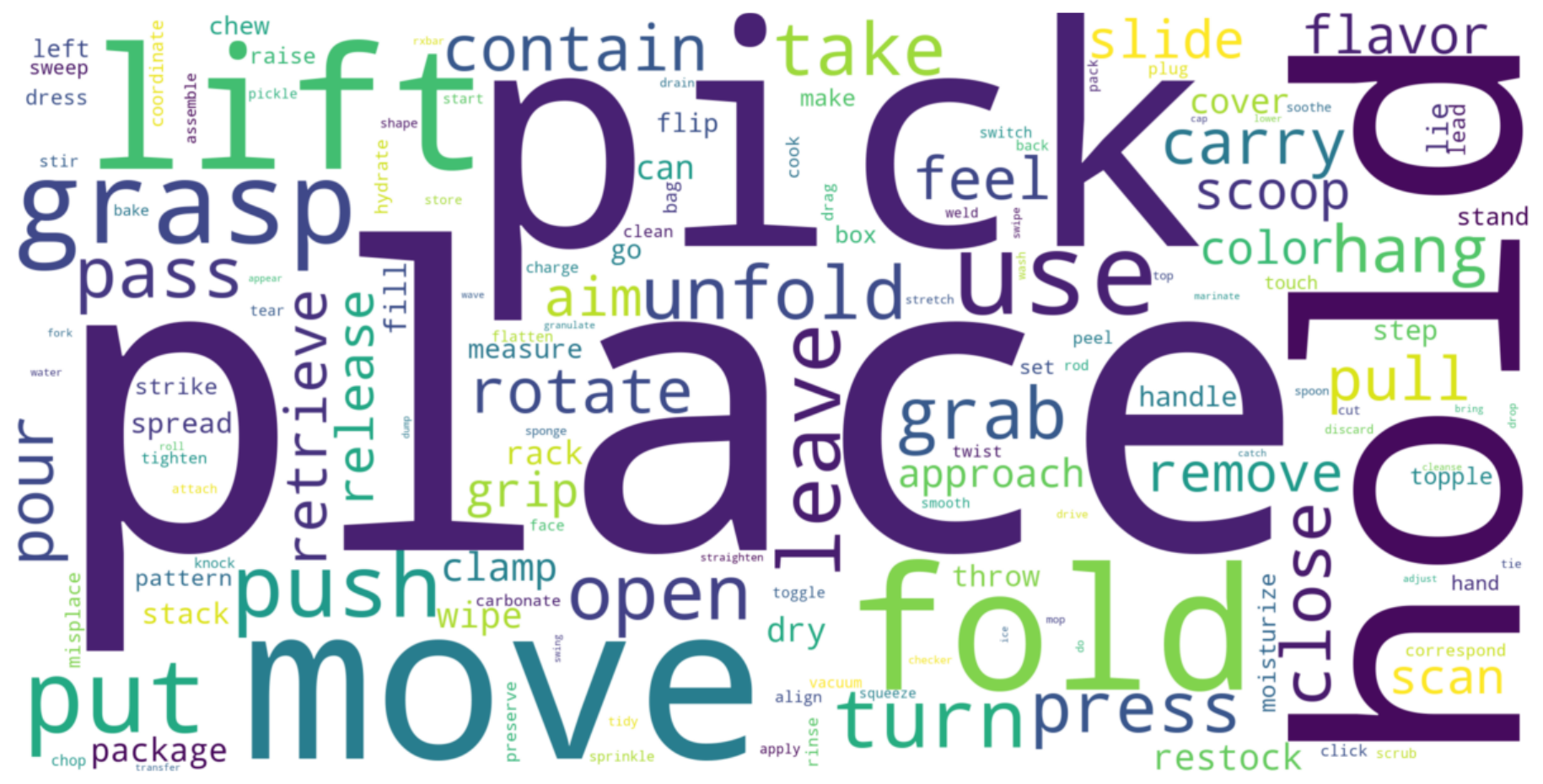}
    \subcaption{Word cloud of task descriptions of existing works.}
  \end{minipage}\hfill
  \begin{minipage}[c]{0.45\linewidth} 
    \centering
    \includegraphics[width=\linewidth]{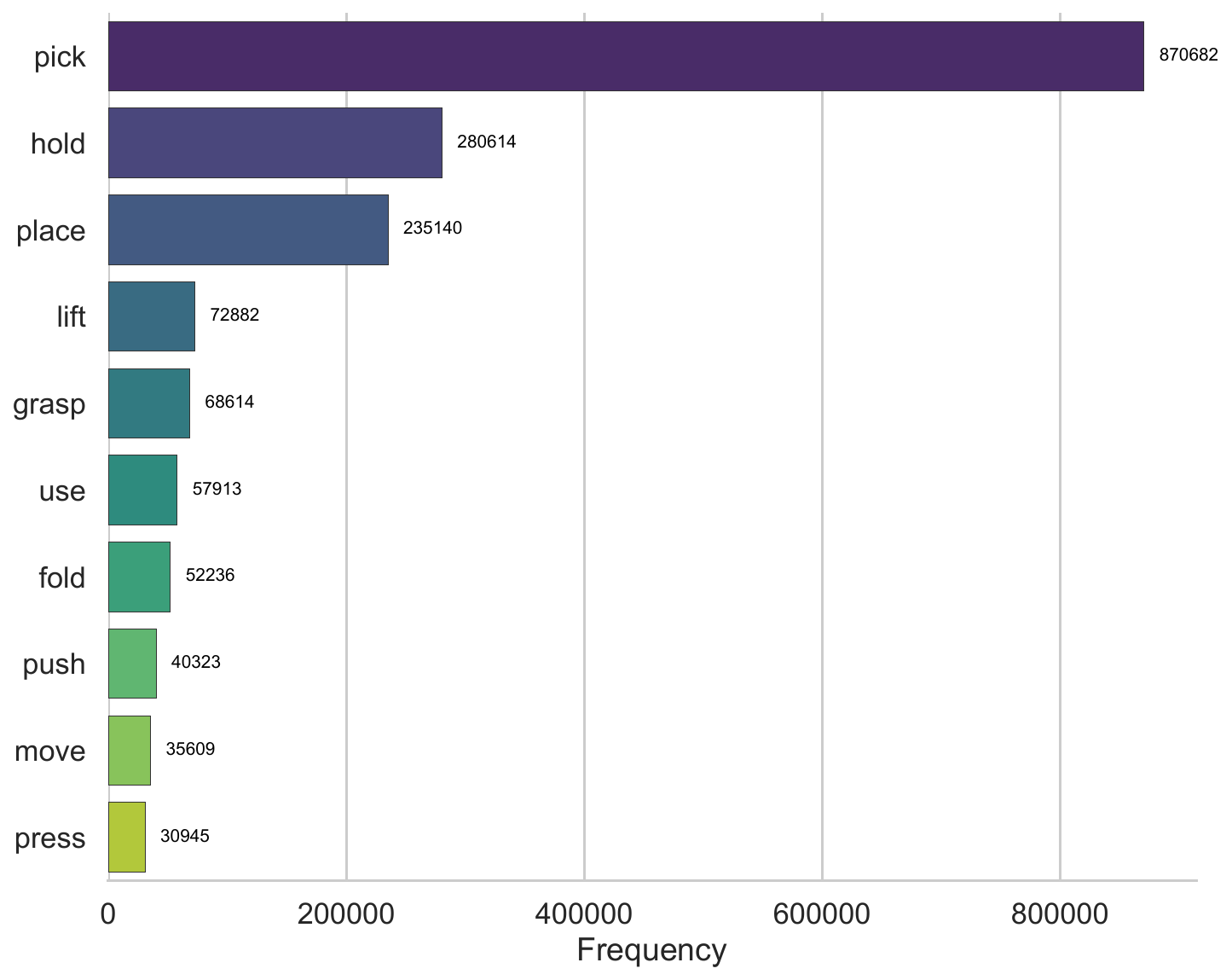}
    \subcaption{Verb frequency distribution in previous task descriptions.}
  \end{minipage}
  
  \caption{Task design analysis from prior works. (a) Word cloud of task descriptions. (b) Verb frequency distribution in task descriptions.}
  \label{fig:fig-task-analysis}
\end{figure}

In previous works, researchers have designed robot tasks based on several subjective criteria, including designer judgment, common daily activities, and application scenarios. 
However, these approaches often lack systematic consideration and design principles, leading to significant overlap in tasks across different works and a focus on very common activities and tasks. This results in insufficient coverage of long-tail tasks in robot datasets and evaluation tasks, with data accumulation concentrated on common tasks while neglecting rare ones.
We collect and analyze the task designs from prior works like Agibot~\cite{bu2025agibot} and Open X-Embodiment~\cite{embodimentcollaboration2025openxembodimentroboticlearning}, removing duplicates and categorizing them. The accompanying word cloud and verb frequency chart in Figure~\ref{fig:fig-task-analysis} reveal a clear bias towards the most common tasks, with many tasks requiring similar actions like ``pick and hold''. 
This analysis highlights the limitations of past task designs, which often overlook rare but important tasks in the long-tail distribution and exhibit significant overlap among different tasks. These issues stem from a longstanding ignorance of the long-tail nature of human activities and the coupling of multi-class actions, which we need robots to learn and perform. 

In light of these, in the work, we propose to design the robot tasks according to the human action propriety. We aim to create a diverse set of tasks that cover a wide range of interactions, considering the coupling of actions and ensuring representation of long-tail, rare but important actions. We also design tasks that may \textit{seem simple in imagination but are actually challenging in practice}, based on insights from human-object interaction studies such as HAKE~\cite{li2019hakehumanactivityknowledge}, PaStaNet~\cite{li2020pastanethumanactivityknowledge}, and OCL~\cite{li2023beyond}.

We first choose the basic interactions based on the previous robot learning works, like the task list from Agibot~\cite{bu2025agibot} and evaluation tasks from $\pi_{0.5}$ \cite{intelligence2025pi_}. In detail, we collect all the tasks from these works, remove duplicates, and categorize them based on their semantic meanings. 
Then, based on these existing tasks, we further expand and supplement the task list by referring to human-object interaction primitives and object affordances from HAKE~\cite{li2019hakehumanactivityknowledge} and OCL~\cite{li2023beyond}.
Using a large language model, we automatically generate a large number of tasks based on carefully designed prompts that incorporate these actions and objects and select a diverse set of activities.

Specifically, we carefully select a set of representative human–object interaction primitives spanning from high-frequency to low-frequency activities. Under a unified task-design prompt, we leverage the Qwen3 model~\cite{qwen3} to automatically generate a large pool of candidate tasks. 
First, we perform word sense disambiguation on the selected action primitives to eliminate potential ambiguities and ensure semantic uniqueness and consistency. We then prompt the Qwen3 to enumerate objects that are semantically and physically relevant to each action primitive. Based on these action–object pairs, the model further synthesizes concrete task instances and refines the task descriptions to produce clear, human-readable textual specifications.

During the task filtering stage, we additionally employ the Qwen3 to automatically score the robot executability of the generated tasks, followed by a final selection conducted by five human experts as the gold standard. 
Through this process, we obtain a high-quality set of robot tasks that are feasible under current hardware constraints and friendly for teleoperation-based data collection.
Finally, we combine evaluations from multiple large language models and human experts to score and select the tasks. We ensure the selected tasks are feasible with current hardware capabilities and are friendly for data collection. We prioritize the tasks based on their scores, and for high-priority tasks, we design specific interaction details and select objects, e.g., choose proper objects from the \url{Taobao.com}. We also establish clear criteria for task completion evaluation, laying the groundwork for future metrics beyond success rate (SR). Additionally, we record template videos of humans completing these tasks to guide data collection.

Based on the tasks generated and selected, and considering the workload for the first version, we select 100 tasks to form the GM-100 benchmark as our 1st version of GM. This set will serve as the foundation for future GM-series research.

\begin{figure}
  \centering
  \includegraphics[width=0.9\linewidth]{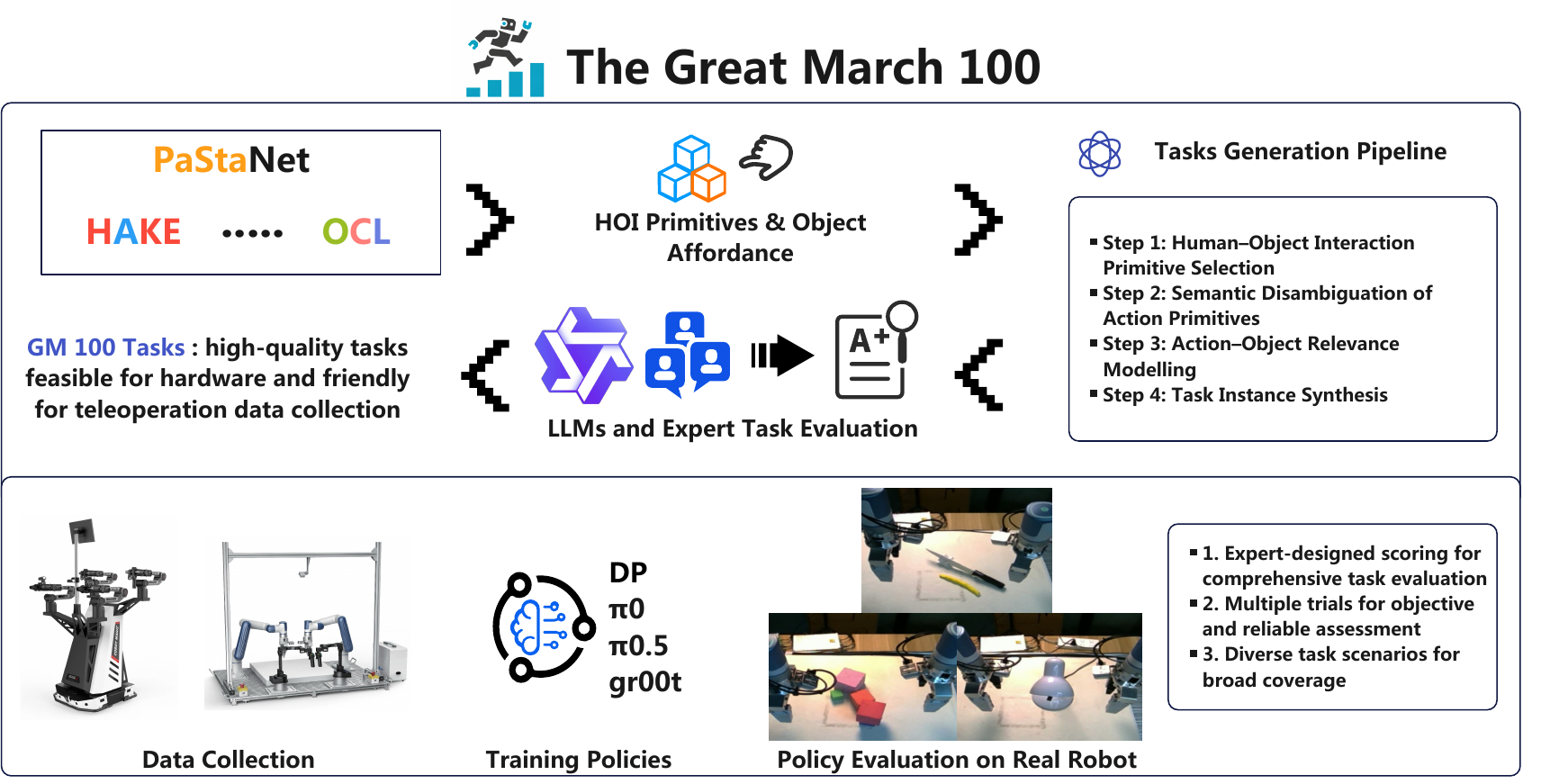}  
  \caption{The construction pipeline of the GM-100 benchmark. The process begins with collecting existing robot tasks, followed by a semantic expansion using HAKE~\cite{li2019hakehumanactivityknowledge} and LLM-based generation to cover long-tail interactions. The candidates then undergo a rigorous hybrid filtration by LLMs and human experts to ensure hardware feasibility and data collection friendliness. Finally, 100 high-priority tasks are selected and instantiated with detailed interaction criteria and template videos.} 
  \label{fig:pipeline}
\end{figure}

\section{GM-100}
\label{sec:dataset}
We collect a medium-sized dataset containing over 13K trajectories with teleoperation in the GM-100 tasks. We open-source the dataset and the task list in \url{https://rhos.ai/research/gm-100}.

\subsection{Hardware Platform}
\begin{figure}
  \centering
  \includegraphics[width=0.8\linewidth]{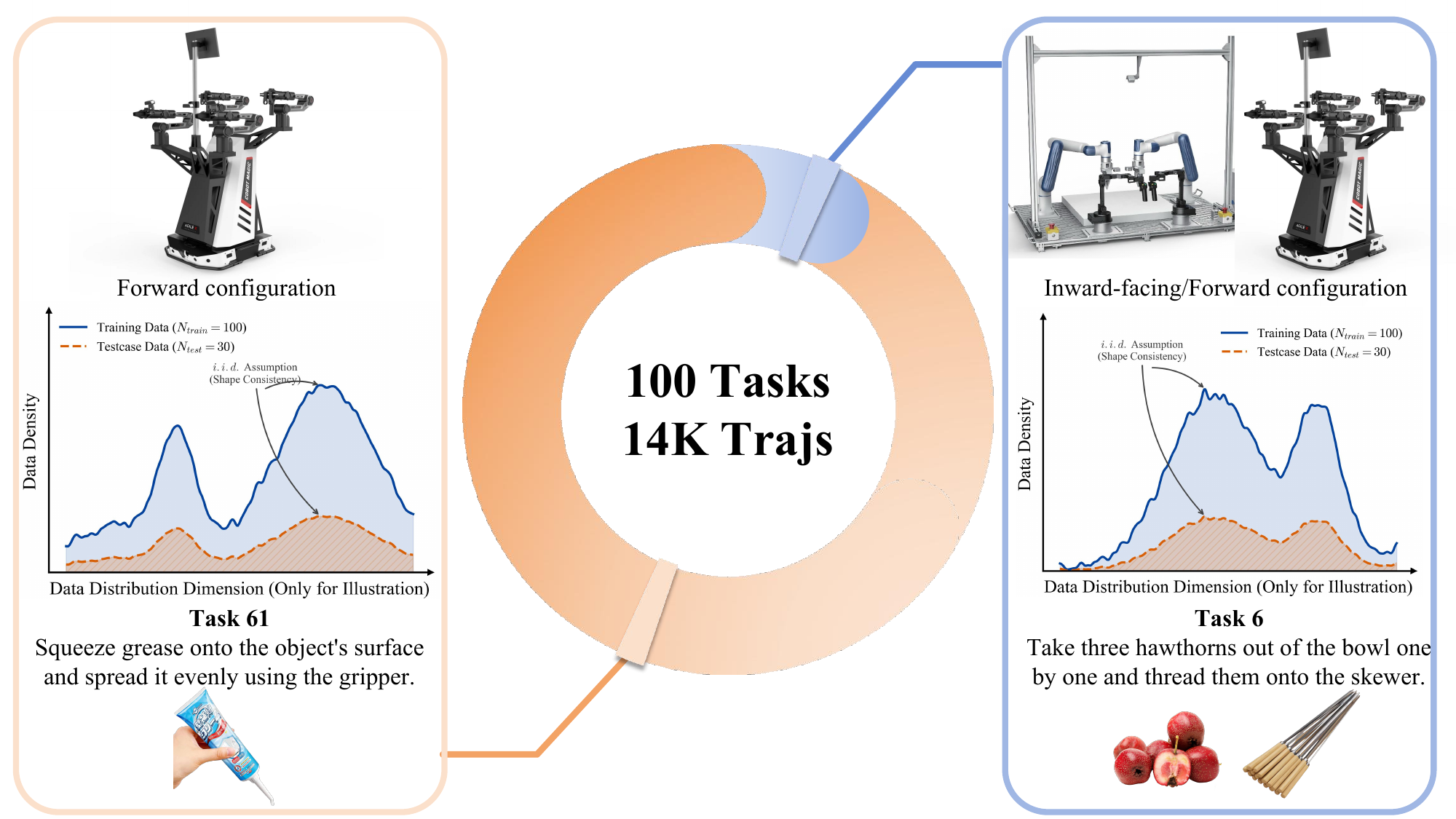}  
  \caption{\textbf{GM-100 Dataset.} Two distinct robotic platforms are utilized for data collection and evaluation. For Tasks 1–10, we collect 130 trajectories per task on both platforms, whereas for Tasks 11–100, data is collected exclusively on the Cobot Magic platform. To ensure the 100 training trajectories and 30 testing trajectories share a similar data distribution, we strive to maintain consistency in the environments and object configurations used for both data collection and evaluation. (Note: The included data distribution figure is schematic and intended solely for illustration purposes; it does not depict the exact statistical distribution.)}
  \label{fig:gm100-dataset}
\end{figure}

We use two robotic platforms, Agilex Cobot Magic and Dobot Xtrainer, to collect the dataset and evaluate the embodied AI agents. The two platforms have different kinematic structures, bi-arm designs, and main camera views, which can provide diverse data for evaluation.
Cobot Magic is a Mobile-Aloha~\cite{fu2024mobile} like robot platform that has a forward-reaching arm structure and a head-mounted camera, while Xtrainer is an Aloha~\cite{zhao2023learning} like platform with an Inward-folding arm structure and a top-down camera view.
The two platforms are shown in Figure~\ref{fig:gm100-dataset}.
The trajectory data is collected by teleoperation, where human operators control the robot arms to perform various tasks. We collect all 100 tasks on Cobot Magic and 10 tasks on Xtrainer for the current version. More data collections are being conducted and will be open-sourced in the next version.

\subsection{Data Distribution}
For each task, we first collect 100 trajectories with different initial conditions and design perturbations to ensure diversity in position, orientation, and object placement.
Then, we collect another 30 trajectories with similar distributions to the first 100 trajectories. These 30 trajectories are used to align the test cases during evaluation, which ensures that the test cases remain consistent across different checkpoints or different models. 

\section{Experimental Setup}
\label{sec:experiment}
\subsection{Baseline Models}
To validate the feasibility and challenge of the GM-100 tasks, we evaluate several baseline models across 100 tasks. The baseline models include DP, $\pi_0$, $\pi_{0.5}$, and GR00T. 
These models are either trained from scratch (for DP) or fine-tuned (for VLA models) on the collected 100 trajectories for each task until convergence. Details of each baseline model and their training procedures are provided in Appendix~\ref{sec:appendix-b}.

\subsection{Evaluation Metrics}
To evaluate the performance of different models on GM-100 tasks, we use the following metrics:
\paragraph{Success Rate (SR).} The percentage of tasks successfully completed within a fixed number of attempts. This is the most commonly used and straightforward metric for evaluating robot task performance. To ensure fairness and reproducibility, we align the test cases across different models by using the same set of 30 test trajectories collected for each task.
Due to the time-consuming nature of real-world robot testing, we will gradually release the results on all baselines over time.

\begin{table}
 \caption{\textbf{Real-world Performance} on Xtrainer platform. Task details are provided in Appendix~\ref{sec:appendix-a}. Due to the time-consuming nature of real-world robot testing, we will gradually release the results on more baselines and tasks over time. Full results can be found at \url{https://rhos.ai/research/gm-100/results}.}
  \label{tab:psr_xtrainer}
  \centering
\begin{tabular}{ccccccc}
\hline
\multirow{2}{*}{Task ID} & \multicolumn{3}{c}{PSR}                     & \multicolumn{3}{c}{SR}                      \\ \cmidrule(r){2-4} \cmidrule(r){5-7} 
                         & DP              & $\pi_0$ & $\pi_{0.5}$     & DP              & $\pi_0$ & $\pi_{0.5}$     \\ \hline
0001                     & 2.5\%           & 45.8\%  & \textbf{72.5\%} & 0.0\%           & 13.3\%  & \textbf{36.7\%} \\
0002                     & 0.5\%           & 35.0\%  & \textbf{39.0\%} & 0.0\%           & 0.0\%   & \textbf{8.0\%}  \\
0003                     & 4.4\%           & 47.8\%  & \textbf{51.1\%} & 0.0\%           & 0.0\%   & 0.0\%           \\
0004                     & 25.8\%          & 45.8\%  & \textbf{70.8\%} & 3.3\%           & 0.0\%   & \textbf{30.0\%} \\
0005                     & \textbf{12.2\%} & 10.2\%  & 8.2\%           & \textbf{12.2\%} & 10.2\%  & 8.2\%           \\
0006                     & 6.3\%           & 19.6\%  & \textbf{60.6\%} & 0.0\%           & 0.0\%   & \textbf{13.3\%} \\
0007                     & 6.2\%           & 44.6\%  & \textbf{90.0\%} & 0.0\%           & 3.3\%   & \textbf{50.0\%} \\
0008                     & 11.1\%          & 42.2\%  & \textbf{78.9\%} & 0.0\%           & 6.7\%   & \textbf{66.7\%} \\
0009                     & 11.1\%          & 20.0\%  & \textbf{31.1\%} & 0.0\%           & 0.0\%   & 0.0\%           \\
0010                     & 0\%             & 10.0\%  & \textbf{36.7\%} & 0.0\%           & 10.0\%  & \textbf{36.7\%} \\ \hline
Average                  & 7.0\%           & 32.1\%  & \textbf{53.9\%} & 1.6\%           & 4.4\%   & \textbf{24.9\%} \\ \hline
\end{tabular}
\end{table}

\paragraph{Partial Success Rate (PSR).} The percentage of subtasks successfully completed within a task. For complex tasks that involve multiple steps or goals, SR alone may not fully capture the model's performance.
Suboptimal robotic arm configurations for specific tasks, the wide distribution of collected datasets, and insufficient training data collectively contribute to low overall success rates on GM-100 benchmarks. 
While this outcome underscores the inherent challenges of the tasks under such data constraints, it also impairs our capacity to conduct fine-grained assessments of model performance.
Thus, for most GM-100 tasks, we define multiple \textbf{subtasks} and \textbf{subgoals} that need to be accomplished to complete the overall task. 
Detailed definitions of score calculation for each task are provided at \href{https://rhos.ai/research/gm-100/tasks}{GM-100 Task List}.
PSR measures how many of these subtasks are successfully achieved, providing a more fine-grained evaluation of model performance.

\paragraph{Action Prediction Error.} The mean squared error (MSE) and L1 loss between the predicted actions and ground truth actions in the specific prediction window on the test trajectories. Although low action prediction error does not guarantee high task success, it reflects the model's ability to understand and replicate unseen expert demonstrations. However, each baseline may predict action chunks of different lengths, so to ensure fair comparison, we compute the action prediction error only on a specific overlapping prediction window across all baselines.

\section{Results and Analysis}
\label{sec:results}
\subsection{Real-world Performance}
\begin{figure}
  \centering
  \includegraphics[width=0.8\linewidth]{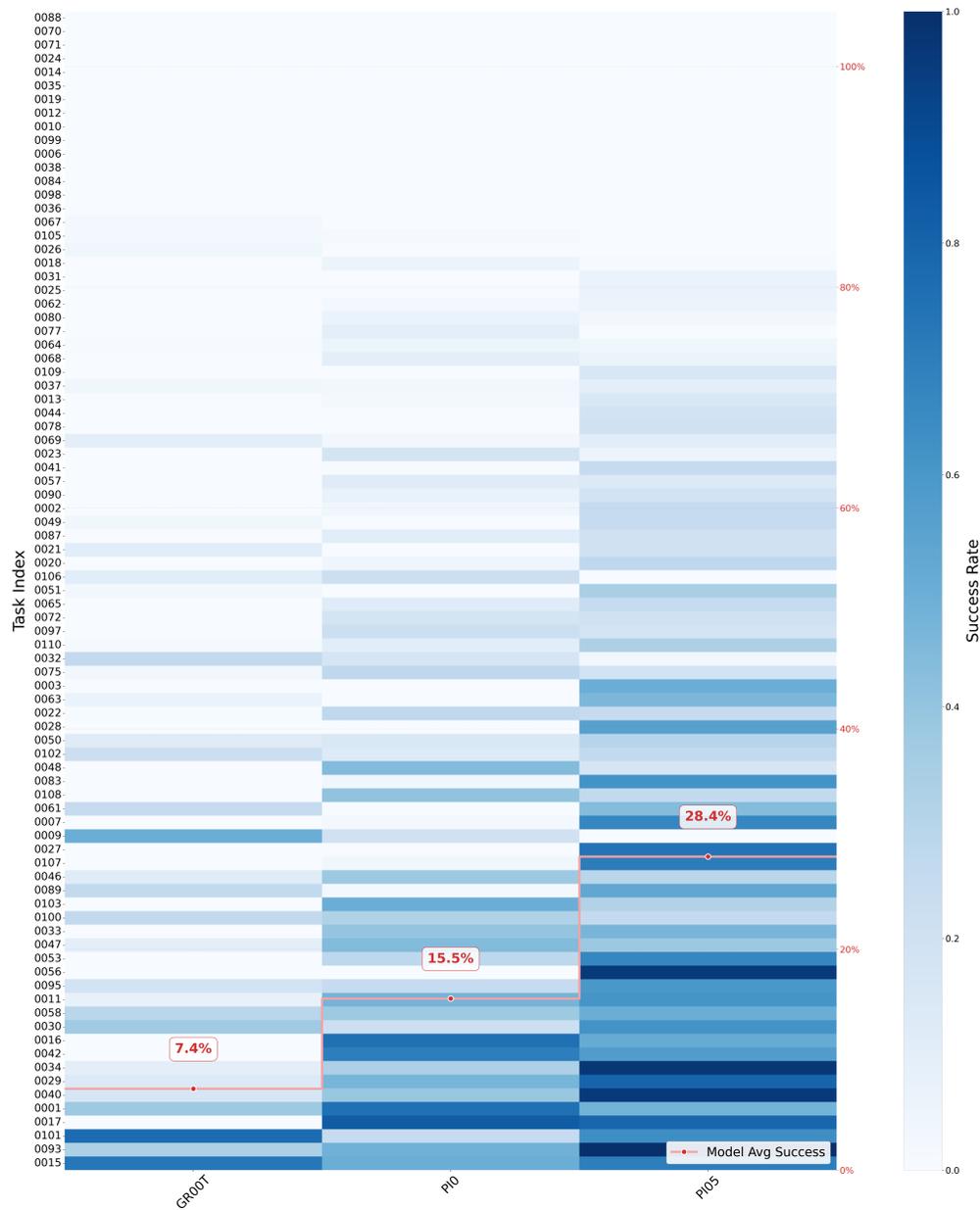}
  \caption{\textbf{Partial Success Rate} on Cobot Magic Platform. 
  The color intensity in the heatmap indicates the PSR. Task Details are provided at \href{https://rhos.ai/research/gm-100/tasks}{GM-100 Task List} due to the space limit. Results on more baselines will be gradually released over time due to the time-consuming nature of real-world robot testing. Detailed PSR can be found at \href{https://rhos.ai/research/gm-100/psr}{GM-100 PSR}.}
  \label{fig:psr_cobot}
\end{figure}

We show the Real-world Performance of different baseline models on the Xtrainer platform in Table~\ref{tab:psr_xtrainer} and Partial Success Rate on the Cobot Magic platform in Figure~\ref{fig:psr_cobot}. Detailed results are provided at \href{https://rhos.ai/research/gm-100/results}{GM-100 Results} due to space limitations.

\subsection{Prediction Loss}
\begin{table}[!htbp]
  \caption{\textbf{Action prediction error analysis} (MSE and L1 loss) on Xtrainer platform. Task Details are provided in Appendix~\ref{sec:appendix-a}.}
  \centering
  \begin{tabular}{lcccccc}
    \toprule
     \multirow{2}{*}{Task ID} & \multicolumn{3}{c}{MSE} & \multicolumn{3}{c}{L1} \\
    \cmidrule(r){2-4} \cmidrule(r){5-7}
    & DP & $\pi_0$ & $\pi_{0.5}$ & DP & $\pi_0$ & $\pi_{0.5}$ \\
    \midrule
    0001 & 0.0041 & 0.0030 & \textbf{0.0025} & 0.0352 & 0.0246 & \textbf{0.0213} \\
    0002 & 0.0030 & 0.0017 & \textbf{0.0015} & 0.0288 & 0.0215 & \textbf{0.0193} \\
    0003 & 0.0036 & 0.0033 & \textbf{0.0027} & 0.0317 & 0.0259 & \textbf{0.0232} \\
    0004 & 0.0062 & 0.0051 & \textbf{0.0045} & 0.0320 & 0.0272 & \textbf{0.0250} \\
    0005 & 0.0038 & \textbf{0.0021} & 0.0022 & 0.0365 & \textbf{0.0215} & 0.0223 \\
    0006 & 0.0028 & 0.0021 & \textbf{0.0019} & 0.0253 & 0.0216 & \textbf{0.0204} \\
    0007 & 0.0101 & 0.0074 & \textbf{0.0064} & 0.0480 & 0.0398 & \textbf{0.0375} \\
    0008 & 0.0010 & 0.0009 & \textbf{0.0007} & 0.0124 & 0.0113 & \textbf{0.0102} \\
    0009 & 0.0091 & 0.0056 & \textbf{0.0042} & 0.0488 & 0.0350 & \textbf{0.0331} \\
    0010 & 0.0031 & 0.0022 & \textbf{0.0020} & 0.0294 & 0.0233 & \textbf{0.0218} \\
    \midrule
    Average & 0.0047 & 0.0033 & \textbf{0.0029} & 0.0328 & 0.0252 & \textbf{0.0234} \\
    \bottomrule
  \end{tabular}
  \label{tab:action_prediction_error}
\end{table}

\begin{figure}[!htbp]
  \centering
  \includegraphics[width=0.7\linewidth]{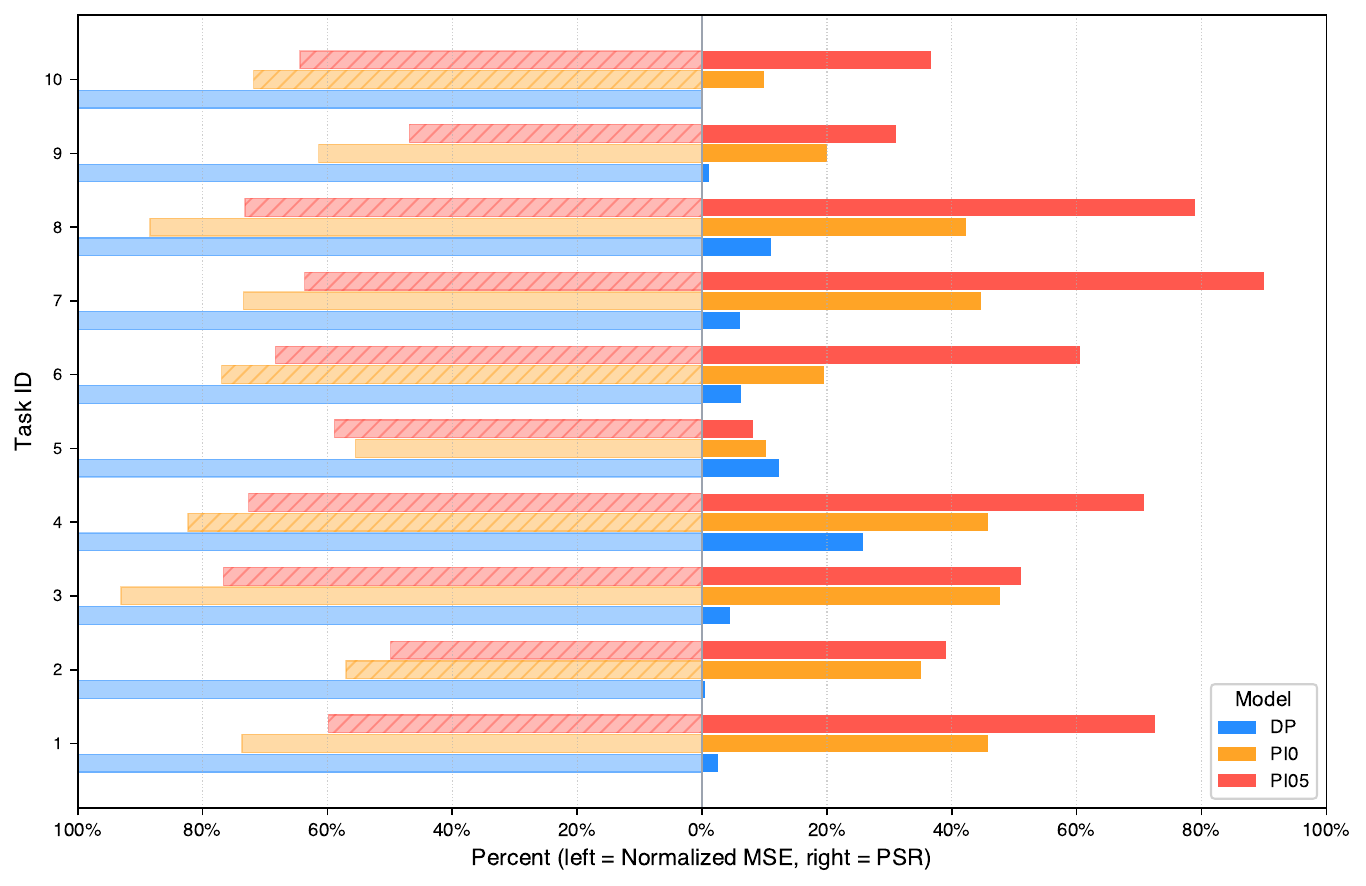}
  \caption{Task-level diverging comparison of \textbf{Normalized MSE} (left) and \textbf{Partial Success Rate} (right) across models.}
  \label{fig:diverging_mse_psr}
\end{figure}

Table~\ref{tab:action_prediction_error} presents the action prediction errors measured by MSE and L1 loss. To further evaluate the relationship between offline supervised objectives and online deployment performance, we visualize the Normalized MSE alongside the Physical Success Rate (PSR) for each task in Figure~\ref{fig:diverging_mse_psr}. 
Here, the Normalized MSE is defined by scaling the raw MSE of each model relative to the maximum MSE recorded among the three candidate models, effectively mapping the error values onto a $[0, 1]$ range to facilitate cross-model comparison. 
A clear inverse correlation is observed between action prediction error and physical success across the three models. Specifically, $\pi_{0.5}$ (represented in red) consistently minimizes the Normalized MSE (leftward bars) while achieving the highest PSR (rightward bars) in nearly all tasks. In contrast, for the Diffusion Policy, a larger deviation in action prediction directly translates to a diminished PSR, which remains below $25\%$ for the majority of tasks. These results suggest that in the Xtrainer environment, high-precision action modeling is a prerequisite for successful physical interaction. The superior performance of $\pi_{0.5}$ indicates that the architectural refinements and training objectives of the $\pi$ series not only minimize the training loss but also effectively capture the underlying distribution of successful trajectories, leading to more robust policy execution in real-world or high-fidelity simulated scenarios.

\section{Website of GM-100}

We have open-sourced our task designs, data, and testing procedures. Since real-world robot testing is highly costly, we will gradually release all results over time.  

We do not aim to build an absolutely fair physical testing environment, as current robotic learning models remain significantly influenced by tester capability and environmental conditions. Achieving truly fair, dispute-free, and mutually trusted global testing may be impractical at this stage. 
Instead, we provide fine-tuning data and task definitions, and operate an open platform where any researchers can upload their own results and evidence videos. 
We will conduct verification and author confirmation to the best of our ability for open-source models. We require weight submissions for review and assign them ``checked'' labels accordingly.  
However, everyone can also update their results without verification, and then the whole community will naturally arrive at a long-term evaluation, as we have done with our papers on arXiv.
We believe in the power of community to foster transparent and credible benchmarking.  

Moving forward, we will continue to expand and refine our task generation system, open-source efforts, and platform maintenance. Our goal is to provide a reliable reference for model evaluation and to build a collaborative open ecosystem around \textbf{GM-X}.  
We welcome partnerships and collaborations with OEMs and model developers. Stay tuned for GM-X!

\section{Conclusion}
In this report, we present the Great March 100 (GM-100), a systematic step towards a comprehensive ``Robot Learning Olympics''. Unlike previous datasets that prioritize scale over structure, GM-100 comprises 100 carefully curated tasks derived from a rigorous analysis of human-object interaction primitives and object affordances. This design philosophy ensures coverage of diverse, long-tail, and rare behaviors that critically test the generalization limits of robotic agents. We don't aim to create just another benchmark; instead, GM-100 serves as a foundational task list for evaluating embodied AI systems in real-world settings.

To support this initiative, we established a medium-sized dataset containing over 13,000 teleoperated trajectories across two distinct robotic platforms: Agilex Cobot Magic and Dobot Xtrainer. Extensive evaluations of baseline models, using metrics such as Success Rate (SR), Partial Success Rate (PSR), and Action Prediction Error, demonstrate that GM-100 tasks are physically feasible yet sufficiently challenging to effectively differentiate the performance of VLAs. Furthermore, acknowledging the challenges of maintaining a ``fair'' physical testing environment, we advocate for a transparent, community-driven evaluation paradigm that relies on collective oversight and open evidence sharing rather than rigid, centralized testing.

Looking forward, GM-100 serves as the foundational layer for GM-X. We are committed to continuously expanding our task generation system and fostering a collaborative, open ecosystem, aiming to build a reliable reference that drives the progress of embodied AI.

\bibliographystyle{unsrt}  
\bibliography{arxiv}  

\appendix
\section{Appendix A: Full Task List}
\label{sec:appendix-a}
Here we provide the details of some GM-100 tasks, along with their descriptions, interaction details, and object purchase links. 
More details aboutthe full 100 tasks, including object specifications and success criteria, can be found at \url{https://rhos.ai/research/gm-100/tasks}.

\keepXColumns 
\renewcommand{\tabularxcolumn}[1]{m{#1}}

\small
\begin{tabularx}{\textwidth}{|c|c|X|X|c|}
    \caption{Robot Manipulation Tasks Dataset} \label{tab:tasks} \\ 
    \hline
    \textbf{Task ID} & \textbf{Image} & \textbf{Description} & \textbf{Interaction Details} & \textbf{Object Links}\\
    \hline
    \endfirsthead 

    \hline
    \textbf{Task ID} & \textbf{Image} & \textbf{Description} & \textbf{Interaction Details} & \textbf{Object Links}\\
    \hline
    \endhead 

    \hline
    \multicolumn{5}{r}{Continued on next page...} \\
    \endfoot 

    \hline
    \endlastfoot 

    0001 & \includegraphics[width=2.5cm, valign=c]{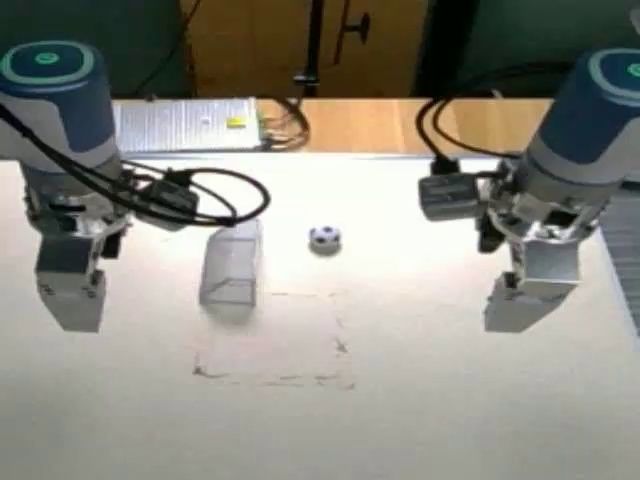} & Use the gripper to strike the small ball into the tabletop goal. & Execute a crisp, instantaneous strike. Do not push. & \href{https://e.tb.cn/h.79AT1BRfHYBB8fW?tk=qTmYUWLil4p}{tabletop ball}\\
    \hline
    0002 & \includegraphics[width=2.5cm, valign=c]{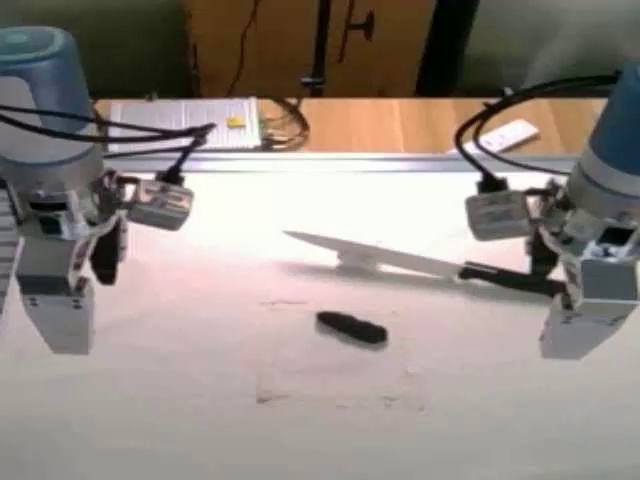} & Grasp the knife from the knife block, slice through clay. & Press blade spine. Use sawing motion if needed. & \href{https://www.printables.com/model/109198-chefs-knife/}{3D knife}\\
    \hline
    0003 & \includegraphics[width=2.5cm, valign=c]{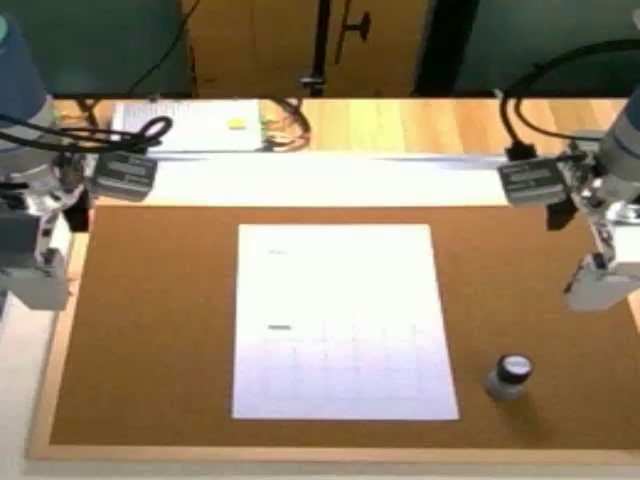} & Grasp the stamp, remove cap, stamp, and re-insert. & Press down firmly and hold for 1-2 seconds. & \href{https://e.tb.cn/h.7kk27lZ0zPoWANo?tk=oTXmUWLtsVb}{stamp}\\
    \hline
    0004 & \includegraphics[width=2.5cm, valign=c]{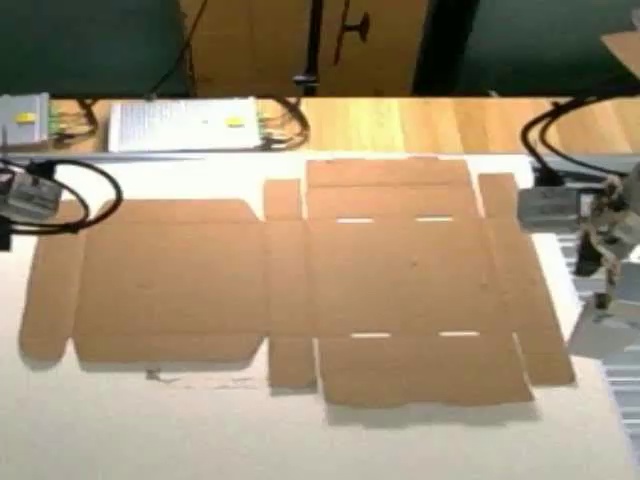} & Fold cardboard template along crease lines. & Only perform a single fold due to constraints. & \href{https://e.tb.cn/h.7kvrJvyQffixx3p?tk=gWQGUWLwsks}{cardboard} \\
    \hline
    0005 & \includegraphics[width=2.5cm, valign=c]{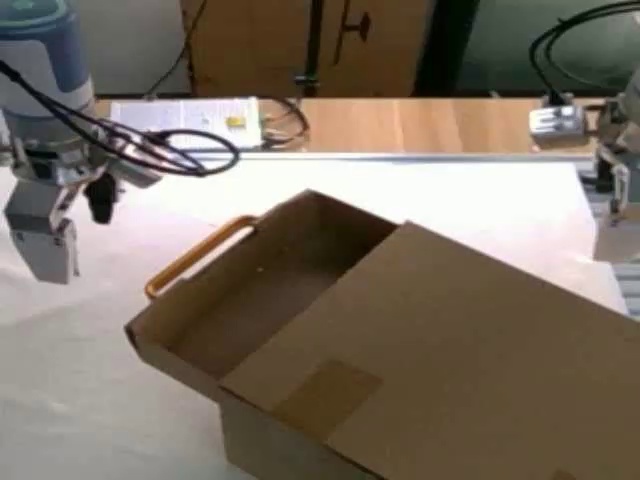} & Close the desktop drawer while stabilizing the rear. & Use second gripper to brace against backward movement. & \href{https://e.tb.cn/h.7l22zJEDrPhd3x3?tk=r9FuUeHl152}{drawer} \\
    \hline
    0006 & \includegraphics[width=2.5cm, valign=c]{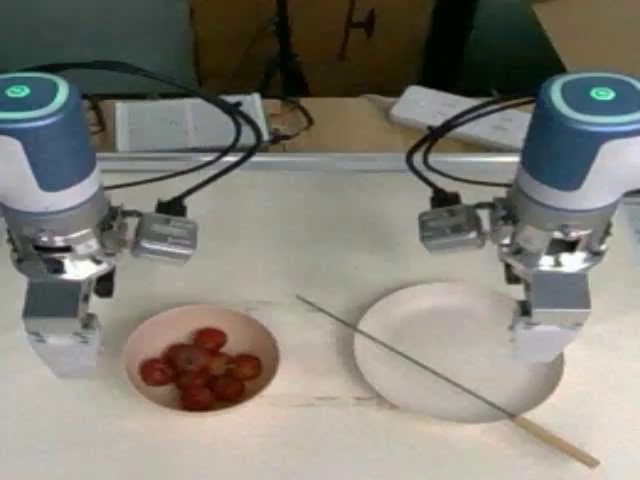} & Thread three hawthorns onto the skewer. & Slide first hawthorns to base to prevent bending. & \thead{\href{https://3.cn/2B0-Khj5?jkl=@EE3LGBarM9@ CA1393}{hawthorn} \\ \href{https://3.cn/2B0KJ-my?jkl=@KCpKbEjMel@}{skewer}}\\
    \hline
    0007 & \includegraphics[width=2.5cm, valign=c]{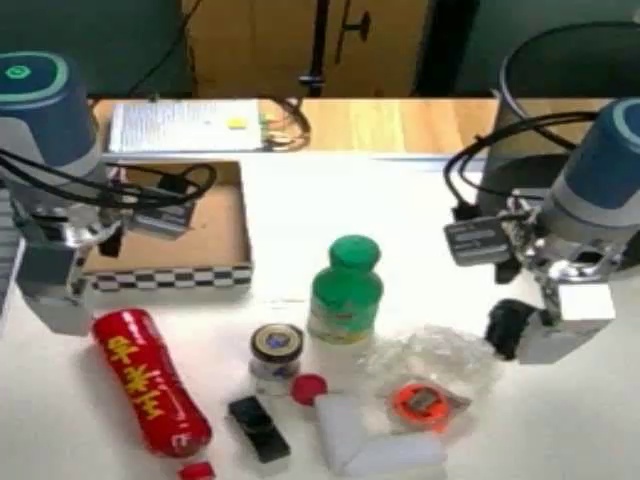} & Clear workspace by tossing trash and placing tools. & Fast and clean tossing motion. Sort trash vs tools. & \href{https://e.tb.cn/h.7lJR3RepJHSobbP?tk=586XUevYIQ1 CA381}{trashbin} \\
    \hline
    0008 & \includegraphics[width=2.5cm, valign=c]{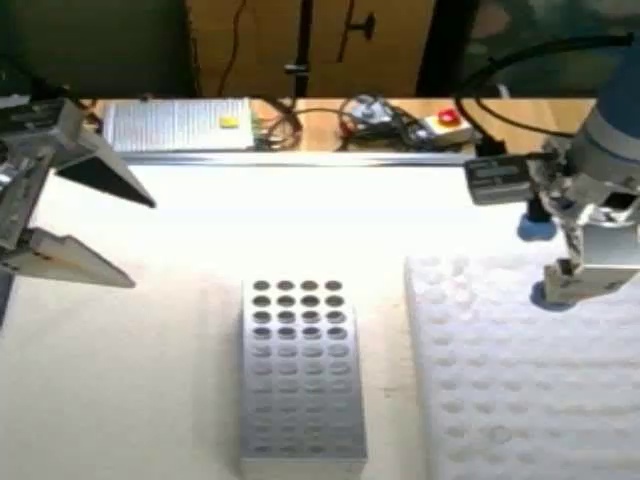} & Transfer test tube from right rack to left rack. & Use compliant grippers to avoid tube failure. &\thead{\href{https://e.tb.cn/h.7l24aN7BlBgIkzK?tk=PNg1UeHo2mb}{tube} \\ \href{https://e.tb.cn/h.7l2epKTWt2d6Jpi?tk=p6uGUeHMeQa}{rack}} \\
    \hline
    0009 & \includegraphics[width=2.5cm, valign=c]{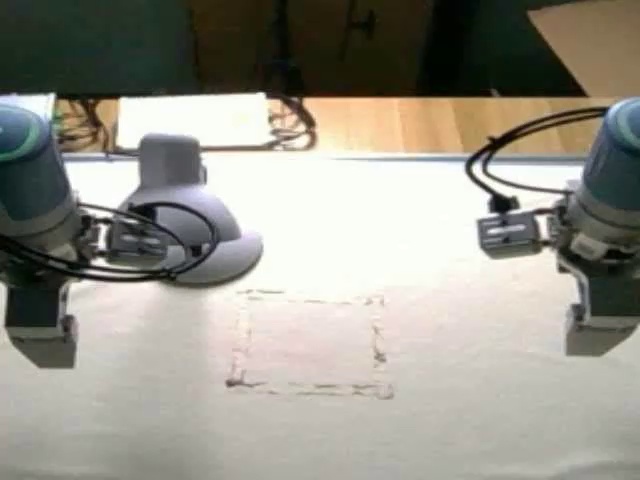} & Press the button of the desk lamp to turn it on. & Use one gripper to hold the lamp, other to press. & \href{https://e.tb.cn/h.7l2QHSCFG2vZ8h4?tk=TBppUeHEorM}{lamp}\\
    \hline
    0010 & \includegraphics[width=2.5cm, valign=c]{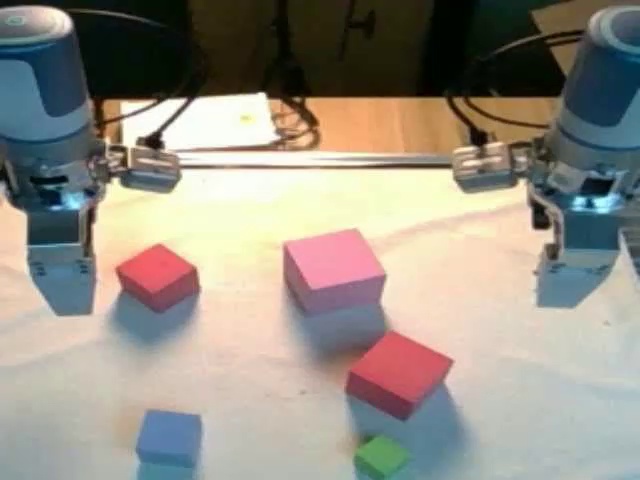} & Sort cubes according to sizes from left to right. & Grasp and compare the sizes of the cubes, placing them from left to right in ascending order while ensuring stable contact with the tabletop, no overlap, and consistent spacing between cubes. & \href{https://e.tb.cn/h.7lAjBVHIjasByxO?tk=V8EsUeHtAJL}{cube}\\
\end{tabularx}

\section{Appendix B: Baseline Model Details}
\label{sec:appendix-b}
Here we provide detailed descriptions of each baseline model used in our experiments, including their architectures, training procedures, and hyperparameter settings.

\paragraph{DP} is finetuned using the LeRobot framework ~\cite{cadene2024lerobot} with a tatal batch size of 512 (8 GPUs $\times$ 64 samples per device) for 100k steps on each task's collected trajectories. The model architecture follows the original design in \cite{chi2023diffusion}, with modifications to accommodate the specific input topic name and action spaces of our robotic platforms. During inference, we utilize an NVIDIA RTX 4090 GPU for real-time action prediction. The inference chunk size is set to 16, and the model executes 8 steps per action prediction cycle.

\paragraph{$\pi_0$ \& $\pi_{0.5}$} is finetuned using the OpenPi framework~\cite{black2024pi_0} with a batch size of 32 for 50k steps on each task's collected trajectories. The model architecture follows the original design in \cite{black2024pi_0}, with modifications to accommodate the specific input topic name and action spaces of our robotic platforms. During inference, we utilize an NVIDIA RTX 4090 GPU for real-time action prediction. The inference chunk size is set to 50, and the model executes 10 steps per action prediction cycle.
\appendix

\end{document}